\documentclass{article}

\usepackage{microtype}
\usepackage{graphicx}
\usepackage{subcaption}
\usepackage{amsmath}
\usepackage{multirow}
\usepackage{xcolor}
\usepackage{booktabs} %

\usepackage[breaklinks=true]{hyperref}

\usepackage[accepted]{sty/icml2021}

\begin{document}

\twocolumn[
\icmltitle{EfficientNetV2: Smaller Models and Faster Training}
\icmlsetsymbol{equal}{*}

\begin{icmlauthorlist}
\icmlauthor{Mingxing Tan}{brain}
\icmlauthor{Quoc V. Le}{brain}
\end{icmlauthorlist}

\icmlaffiliation{brain}{Google Research, Brain Team}

\icmlcorrespondingauthor{Mingxing Tan}{tanmingxing@google.com}

\icmlkeywords{Machine Learning, ICML}

\vskip 0.3in
]

\printAffiliationsAndNotice{}  %

\def\xnet{EfficientNetV2}
\def\TODO{\textcolor{red}{\emph{TODO: }}}
\newcommand{\TT}[1]{\texttt{#1}}
\newcommand{\BF}[1]{\textbf{#1}}
\newcommand{\IT}[1]{\textit{#1}}

\newcommand\blfootnote[1]{%
	\begingroup
	\renewcommand\thefootnote{}\footnote{#1}%
	\addtocounter{footnote}{-1}%
	\endgroup
}

\newcommand{\good}[1]{{\small\color{blue}(#1)}}                                      
\newcommand{\bad}[1]{{\small\color{red}(#1)}}
\newcommand{\std}[1]{{\footnotesize \textcolor{gray}{$\pm$#1}}}
\newcommand{\bd}[1]{\textbf{#1}}                                                    
\newcommand{\app}{\raise.17ex\hbox{$\scriptstyle\sim$}}                             
\newcommand{\symb}[1]{{\small\texttt{#1}}\xspace}                                   
\newcommand{\mrtwo}[1]{\multirow{2}{*}{#1}}                                         
\def\x{\times}                                                                      
\def\pt{p_\textrm{t}}                                                               
\def\at{\alpha_\textrm{t}}                                                          
\def\xt{x_\textrm{t}}                                                               
\def\CE{\textrm{CE}}                                                                
\def\FL{\textrm{FL}}                                                                
\def\FQ{\textrm{FL}^*}                                                              
\newcommand{\eqnnm}[2]{\begin{equation}\label{eq:#1}#2\end{equation}\ignorespaces}

\newlength\savewidth\newcommand\shline{\noalign{\global\savewidth\arrayrulewidth 
		\global\arrayrulewidth 1pt}\hline\noalign{\global\arrayrulewidth\savewidth}}   
\newcommand{\tablestyle}[2]{\setlength{\tabcolsep}{#1}\renewcommand{\arraystretch}{#2}\centering\footnotesize}
\makeatletter\renewcommand\paragraph{\@startsection{paragraph}{4}{\z@}              
	{.5em \@plus1ex \@minus.2ex}{-.5em}{\normalfont\normalsize\bfseries}}\makeatother
\renewcommand{\dbltopfraction}{1}                                                   
\renewcommand{\bottomfraction}{0}                                                   
\renewcommand{\textfraction}{0}                                                     
\renewcommand{\dblfloatpagefraction}{0.95}                                          
\setcounter{dbltopnumber}{5}

\newcommand{\M}[1]{\mathcal{#1}}
\begin{abstract}
\label{sec:abstract}
\setcounter{footnote}{1}

This paper introduces {\xnet}, a new family of convolutional networks that have faster training speed and better parameter efficiency than previous models. 
To develop these models, we use a combination of training-aware neural architecture search and scaling, to jointly optimize training speed and parameter efficiency. The models were searched from the search space enriched with new ops such as Fused-MBConv.
Our experiments show that EfficientNetV2 models train much faster than state-of-the-art models while being up to 6.8x smaller.

Our training can be further sped up by progressively increasing the image size during training, but it often causes a drop in accuracy.  To compensate for this accuracy drop, we propose an improved method of progressive learning, which adaptively adjusts regularization (e.g. data augmentation) along with image size.

With progressive learning, our {\xnet} significantly outperforms previous models on ImageNet and CIFAR/Cars/Flowers datasets. 
By pretraining on the same ImageNet21k, our {\xnet} achieves 87.3\% top-1 accuracy on ImageNet ILSVRC2012, outperforming the recent ViT by 2.0\% accuracy  while training 5x-11x faster using the same computing resources. Code is available at  {\small \url{https://github.com/google/automl/tree/master/efficientnetv2}}.

\end{abstract}
\section{Introduction}

\begin{figure}[!t]
	\centering
	\includegraphics[width=0.99\columnwidth]{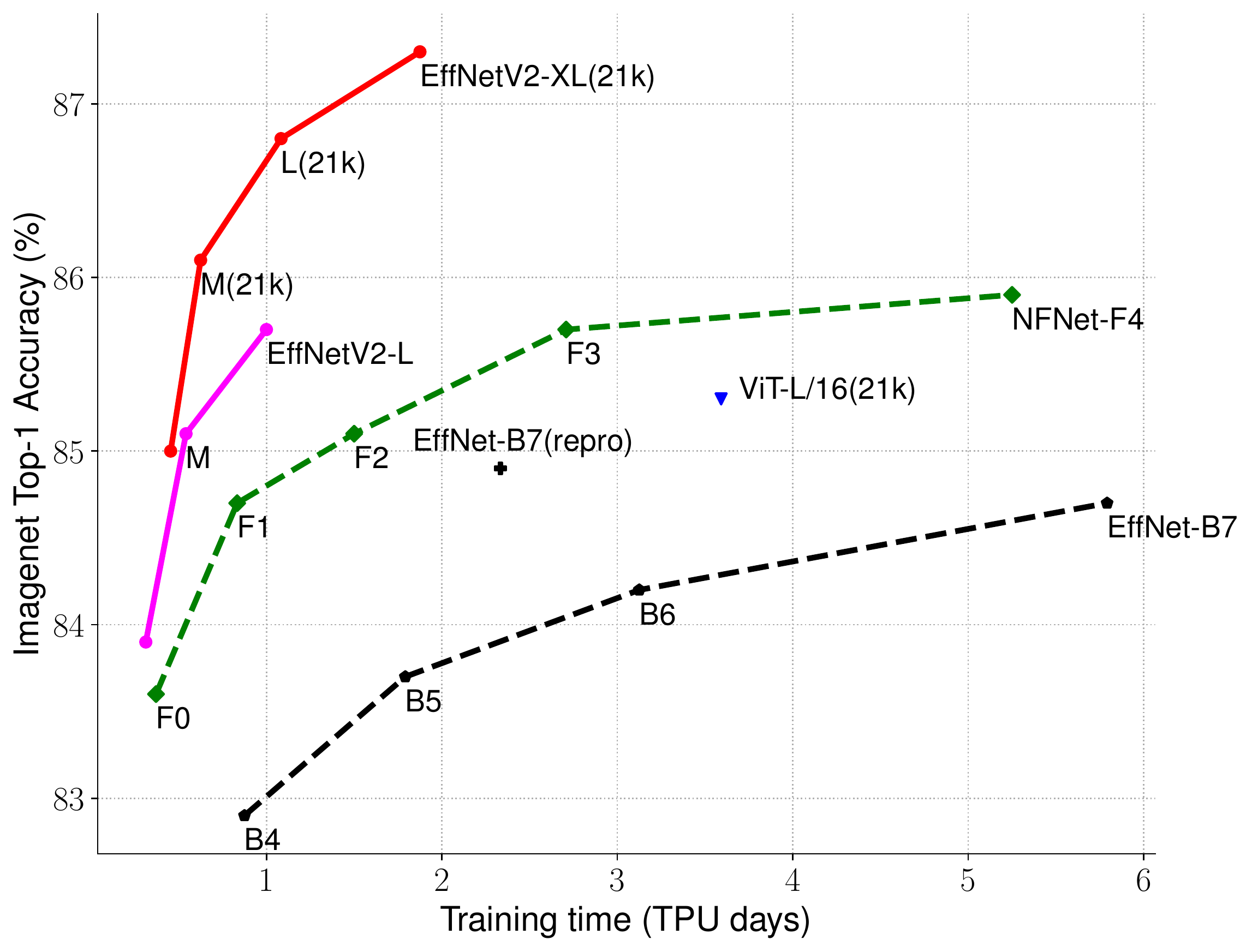}
	\vskip -0.05in
    (a) Training efficiency.
	\vskip 0.1in
	\resizebox{0.99\columnwidth}{!}{
		\begin{tabular}{c|cccc}
			\toprule
						&   EfficientNet & ResNet-RS & DeiT/ViT & {\xnet} \\
						&   (2019)   & (2021)    & (2021) & (ours) \\
			\midrule
			Top-1 Acc.  &   84.3\%   & 84.0\%   &  83.1\%   & 83.9\% \\
			Parameters      &    43M     & 164M   &  86M   & 24M  \\
			\bottomrule
		\end{tabular}
	}
    (b) Parameter efficiency.
    \caption{
		\BF{ImageNet ILSVRC2012 top-1 Accuracy vs. Training Time and Parameters}  -- Models tagged with \TT{21k} are pretrained on ImageNet21k, and others are directly trained on ImageNet ILSVRC2012. Training time is measured with 32 TPU cores.
		All {\xnet} models are trained with progressive learning.
		Our {\xnet} trains 5x - 11x faster than others, while using up to 6.8x fewer parameters.
		Details are in Table \ref{tab:imagenet} and Figure \ref{fig:infer2acc}.
    }
	\label{fig:train2acc}
	\vskip -0.1in
\end{figure}
 Training efficiency is important to deep learning as model size and training data size are increasingly larger. For example, GPT-3 \cite{gpt320},  with much a larger model and more training data,  demonstrates the remarkable capability in few shot learning,
but it requires weeks of training with thousands of GPUs, making it  difficult  to retrain or improve. 

Training efficiency has gained significant interests recently. For instance, NFNets~\cite{nfnet21} aim to improve training efficiency by removing the expensive batch normalization; Several recent works~\cite{botnet21} focus on improving training speed by adding attention layers into convolutional networks (ConvNets); Vision Transformers~\cite{vit21} improves training efficiency on large-scale datasets by using Transformer blocks. However, these methods often come with expensive overhead on large parameter size, as shown in Figure 1(b).%

In this paper, we use an combination of training-aware neural architecture search (NAS) and scaling  to improve both training speed and parameter efficiency. Given the parameter efficiency of EfficientNets~\cite{efficientnet19}, we start by systematically studying the training bottlenecks in EfficientNets. Our study shows in EfficientNets: (1) training with very large image sizes is slow; (2) depthwise convolutions are slow in early layers. (3) equally scaling up every stage is sub-optimal. Based on these observations, we design a search space enriched with additional ops such as Fused-MBConv, and apply training-aware NAS and scaling to jointly optimize model accuracy, training speed, and parameter size. Our found networks, named \emph{{\xnet}}, train up to 4x faster than prior models (Figure \ref{fig:steptime}), while being up to 6.8x smaller in parameter size.

Our training can be further sped up by progressively increasing image size during training. Many previous works, such as progressive resizing~\cite{fastaidawnbench}, FixRes~\cite{fixres20}, and Mix\&Match~\cite{mixmatch19}, have used smaller image sizes in training; however, they usually keep the same regularization for all image sizes, causing a drop in accuracy. We argue that keeping the same regularization for different image sizes is not ideal: for the same network, small image size leads to small network capacity and thus requires weak
regularization; vice versa, large image size requires stronger regularization to combat overfitting (see Section \ref{sec:motivation}). Based on this insight, we propose an improved method of \emph{progressive learning}: in the early training epochs, we train the network with small image size and weak regularization (e.g., dropout and data augmentation), then we gradually increase  image size and add stronger regularization. Built upon progressive resizing~\cite{fastaidawnbench}, but by dynamically adjusting regularization, our approach can speed up the training without causing accuracy drop.

With the improved progressive learning, our {\xnet} achieves 
strong results  on ImageNet, CIFAR-10, CIFAR-100, Cars, and Flowers dataset. On ImageNet, we achieve 85.7\% top-1 accuracy while training 3x - 9x faster and being up to 6.8x smaller than previous models (Figure \ref{fig:train2acc}). 
Our {\xnet} and progressive learning also make it easier to train models on larger datasets. For example, ImageNet21k ~\cite{imagenet15} is about 10x larger than ImageNet ILSVRC2012, but our {\xnet} can finish the training within two days using moderate computing resources of 32 TPUv3 cores. By pretraining on the public ImageNet21k, our {\xnet} achieves 87.3\% top-1 accuracy on ImageNet ILSVRC2012, outperforming the recent ViT-L/16 by 2.0\% accuracy while training 5x-11x faster (Figure \ref{fig:train2acc}).

Our contributions are threefold: 
\begin{itemize}
    \item We introduce {\xnet}, a new family of smaller and faster models. Found by our training-aware NAS and scaling,  {\xnet} outperform previous models in both training speed and parameter efficiency.

    \item We propose an improved method of progressive learning, which adaptively adjusts regularization along with image size. We show that it speeds up training, and simultaneously improves accuracy.

    \item We demonstrate up to 11x faster training speed and up to 6.8x better parameter efficiency on ImageNet, CIFAR, Cars, and Flowers dataset, than prior art.
\end{itemize}
\section{Related work}

\paragraph{Training and parameter efficiency:} Many works, such as DenseNet~\cite{densenet17} and EfficientNet~\cite{efficientnet19}, focus on parameter efficiency, aiming to achieve better accuracy with less parameters. Some more recent works aim to improve training or inference speed instead of parameter efficiency. For example, RegNet~\cite{regnet20}, ResNeSt~\cite{resnest20}, TResNet~\cite{tresnet20}, and EfficientNet-X~\cite{efficientnetx21} focus on GPU and/or TPU inference speed; NFNets~\cite{nfnet21} and BoTNets~\cite{botnet21} focus on improving training speed. However, their training or inference speed often comes with the cost of more parameters. This paper aims to significantly improve both training speed and parameter efficiency than prior art.\looseness=-1 

\paragraph{Progressive training:} Previous works have proposed different kinds of progressive training, which dynamically change the training settings or networks, for GANs~\cite{progressivegan18}, transfer learning~\cite{progressivegan18}, adversarial learning~\cite{progressiveadv19}, %
and language models~\cite{shortformer21}. Progressive resizing~\cite{fastaidawnbench} is mostly related to our approach, which aims to improve training speed. However, it usually comes with the cost of accuracy drop. Another closely related work is Mix\&Match~\cite{mixmatch19}, which randomly sample different image size for each batch. Both progressive resizing and Mix\&Match use the same regularization for all image sizes, causing a drop in accuracy. In this paper, our main difference is to adaptively adjust regularization as well so that we can improve both training speed and accuracy. Our approach is also partially inspired by curriculum learning~\cite{curriculum09}, which schedules training examples from easy to hard. Our approach also gradually increases learning difficulty by adding more regularization, but we don't selectively pick training examples.

\paragraph{Neural architecture search (NAS):}
By automating the network design process, NAS has been used to optimize the network architecture for image classification~\cite{nas_imagenet18}, object detection~\cite{detnas19,efficientdet20}, segmentation~\cite{autodeeplab19}, hyperparameters~\cite{autohas20}, and other applications~\cite{elsken2019neural}. Previous NAS works mostly focus on improving FLOPs efficiency~\cite{mixconv19,efficientnet19} or inference efficiency~\cite{mnas19,proxyless19,fbnet19,efficientnetx21}. Unlike prior works, this paper uses NAS to optimize training and parameter efficiency.  %
\section{{\xnet} Architecture Design}
\label{sec:xnet}

In this section, we study the training bottlenecks of EfficientNet~\cite{efficientnet19}, and introduce our training-aware NAS and scaling, as well as {\xnet} models.

\subsection{Review of EfficientNet}
EfficientNet~\cite{efficientnet19} is a family of models that are optimized for FLOPs and parameter efficiency. It leverages NAS to search for the baseline EfficientNet-B0 that has better trade-off on accuracy and FLOPs. The baseline model is then scaled up with a compound scaling strategy to obtain a family of models B1-B7.
While recent works have claimed large gains on training or inference speed, they are often worse than EfficientNet in terms of parameters and FLOPs efficiency (Table \ref{tab:paramscompare}). 
In this paper, we aim to improve the training speed while  maintaining the parameter efficiency.

\begin{table}[h]
    \vskip -0.15in
    \centering
    \caption{EfficientNets have good parameter and FLOPs efficiency.}
    \resizebox{0.99\columnwidth}{!}{
     \begin{tabular}{crrr}
        \toprule
                            & Top-1 Acc.  & Params    & FLOPs  \\
        \midrule
        EfficientNet-B6~\cite{efficientnet19} &  84.6\%  &  \BF{43M}   &  \BF{19B} \\
        ResNet-RS-420~\cite{resnetrs21}    &      84.4\%  &  192M  & 64B \\
        NFNet-F1~\cite{nfnet21} &                84.7\%  &  133M  &  36B \\
        \bottomrule
     \end{tabular}
    }
    \vskip -0.1in
\label{tab:paramscompare}
\end{table} 

\subsection{Understanding Training Efficiency}

We study the training bottlenecks of EfficientNet~\cite{efficientnet19}, henceforth is also called EfficientNetV1, and a few simple techniques to improve training speed.

\paragraph{Training with very large image sizes is slow:}
As pointed out by previous works~\cite{regnet20}, EfficientNet's large image size results in significant memory usage. Since the total memory on GPU/TPU is fixed, we have to train these models with smaller batch size, which drastically slows down the training. A simple improvement is to apply FixRes~\cite{fixres20}, by using a smaller image size for training than for inference. As shown in Table \ref{tab:fixres}, smaller image size leads to less computations and enables large batch size,  and thus improves training speed by up to 2.2x. Notably, as pointed out in \cite{fixefficientnet20,nfnet21}, using smaller image size for training also leads to slightly better accuracy. But unlike \cite{fixres20}, we do not finetune any layers after training.

\begin{table}[h]
    \vskip -0.1in
    \centering
    \caption{EfficientNet-B6 accuracy and training throughput for different batch sizes and image size.}
    \resizebox{0.99\columnwidth}{!}{
     \begin{tabular}{cr|rr|rr}
        \toprule
                       &          &  \multicolumn{2}{c|}{TPUv3 imgs/sec/core} & \multicolumn{2}{c}{V100 imgs/sec/gpu}  \\
                       & Top-1 Acc.  & batch=32 & batch=128 &batch=12 & batch=24  \\
        \midrule
        train size=512 & 84.3\%  &  42  & OOM & 29 & OOM \\
        train size=380 & 84.6\%  &  76  & 93  & 37 & 52 \\
        \bottomrule
     \end{tabular}
    }
    \vskip -0.1in
\label{tab:fixres}
\end{table} 
In Section \ref{sec:plearn}, we will explore a more advanced training approach, by progressively adjusting image size and regularization during training. 

\begin{figure}[h]
    \centering
    \includegraphics[width=0.6\linewidth]{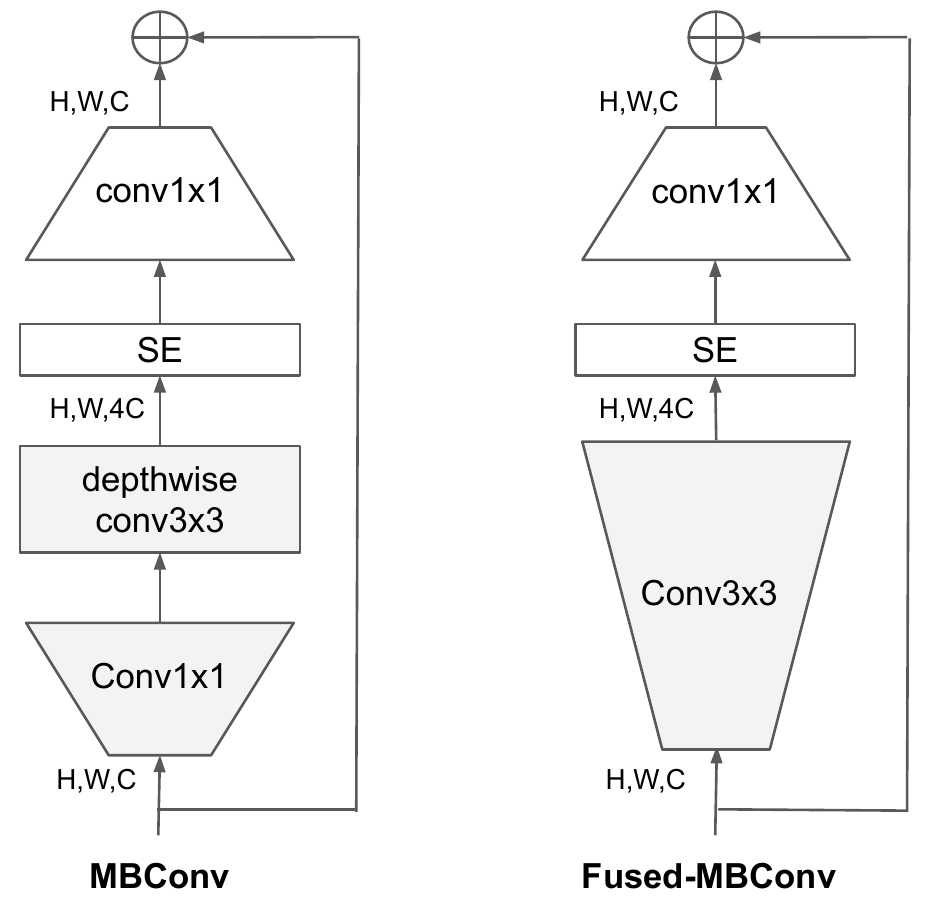}
    \vskip -0.1in
    \caption{Structure of MBConv and Fused-MBConv.}
    \label{fig:fusedmbconv}
    \vskip -0.1in
\end{figure}

\paragraph{Depthwise convolutions are slow in early layers but effective in later stages:}
Another training bottleneck of EfficientNet comes from the extensive depthwise convolutions~\cite{sepconv14}. Depthwise convolutions have fewer parameters and FLOPs than regular convolutions, but they often cannot fully utilize modern accelerators. Recently, Fused-MBConv is proposed in \cite{efficientnetedgetpu19} and later used in \cite{automledgetpu20,mobiledet20,efficientnetx21} to better utilize mobile or server accelerators. It replaces the depthwise conv3x3 and expansion conv1x1 in MBConv~\cite{mobilenetv218,efficientnet19} with a single regular conv3x3, as shown in Figure \ref{fig:fusedmbconv}. To systematically compares these two building blocks, we gradually replace the original MBConv in EfficientNet-B4 with Fused-MBConv (Table \ref{tab:fusedconv}). When applied in early stage 1-3, Fused-MBConv can improve training speed with a small overhead on parameters and FLOPs, but if we replace all blocks with Fused-MBConv (stage 1-7), then it significantly increases parameters and FLOPs while also slowing down the training. Finding the right combination of these two building blocks, MBConv and Fused-MBConv, is non-trivial, which motivates us to leverage neural architecture search to automatically search for the best combination.

\begin{table}[!ht]
    \vskip -0.1in
    \caption{Replacing MBConv with Fused-MBConv. \TT{No fused} denotes all stages use MBConv, \TT{Fused stage1-3} denotes replacing MBConv with Fused-MBConv in stage \{2, 3, 4\}.
    }
    \centering
    \resizebox{0.99\columnwidth}{!}{
        \begin{tabular}{l|rrr|cc}
        \toprule  
                         &  Params  & FLOPs & Top-1  & TPU  & V100 \\
                         &  (M)  & (B)  & Acc.  & imgs/sec/core  & imgs/sec/gpu \\
        \midrule
        No fused         &   19.3   & 4.5   &   82.8\% & 262 & 155 \\
        Fused stage1-3  &    20.0   & 7.5   &   83.1\% & 362 & 216 \\
        Fused stage1-5  &    43.4   & 21.3  &  83.1\%  & 327 & 223 \\
        Fused stage1-7  &    132.0  & 34.4  &  81.7\%  & 254 & 206 \\
        \bottomrule                                                      
    \end{tabular}
    }
\label{tab:fusedconv} 
\end{table} 

\paragraph{Equally scaling up every stage is sub-optimal:} 
EfficientNet equally scales up all stages using a simple compound scaling rule. For example, when depth coefficient is 2, then all stages in the networks would double the number of layers. However, these stages are not equally contributed to the training speed and parameter efficiency. In this paper, we will use a non-uniform scaling strategy to gradually add more layers to later stages. In addition, EfficientNets aggressively scale up image size, leading to large memory consumption and slow training. To address this issue, we slightly modify the scaling rule and restrict the maximum image size to a smaller value.

\subsection{Training-Aware NAS and Scaling}

To this end, we have learned multiple design choices for improving training speed. To search for the best combinations of those choices, we now propose a training-aware NAS.

\paragraph{NAS Search:} Our training-aware NAS framework is largely based on previous NAS works~\cite{mnas19,efficientnet19}, but aims to jointly optimize accuracy, parameter efficiency, and training efficiency on modern accelerators. Specifically, we use EfficientNet as our backbone. Our search space is a stage-based factorized space similar to~\cite{mnas19}, which consists of the design choices for convolutional operation types \{MBConv, Fused-MBConv\}, number of layers, kernel size \{3x3, 5x5\}, expansion ratio \{1, 4, 6\}. On the other hand, we reduce the search space size by (1) removing unnecessary search options such as pooling skip ops, since they are never used in the original EfficientNets; (2) reusing the same channel sizes from the backbone as they are already searched in~\cite{efficientnet19}. Since the search space is smaller, we can apply reinforcement learning~\cite{mnas19} or simply random search on much larger networks that have comparable size as EfficientNet-B4. Specifically, we sample up to 1000 models and train each model about 10 epochs with reduced image size for training. Our search reward combines the model accuracy $A$, the normalized training step time $S$, and the parameter size $P$, using a simple weighted product $A\cdot S^w \cdot P^v$,  where $w$ = -0.07 and $v$ = -0.05 are empirically determined to balance the trade-offs similar to~\cite{mnas19}.

\paragraph{{\xnet} Architecture:} Table \ref{tab:arch} shows the architecture for our searched model {\xnet}-S. Compared to the EfficientNet backbone, our searched {\xnet} has several major distinctions: (1) The first difference is {\xnet} extensively uses both MBConv~\cite{mobilenetv218,efficientnet19} and the newly added fused-MBConv~\cite{efficientnetedgetpu19} in the early layers. (2) Secondly, {\xnet} prefers smaller expansion ratio for MBConv since smaller expansion ratios tend to have less memory access overhead. (3) Thirdly, {\xnet} prefers smaller 3x3 kernel sizes, but it adds more layers to compensate the reduced receptive field resulted from the smaller kernel size. (4) Lastly, {\xnet} completely removes the last stride-1 stage in the original EfficientNet, perhaps due to its large parameter size and memory access overhead.\looseness=-1

\begin{table}[t]
    \caption{
        {\xnet}-S architecture -- MBConv and Fused-MBConv blocks are described in Figure \ref{fig:fusedmbconv}.
    }
    \centering

    \resizebox{0.95\columnwidth}{!}{
        \begin{tabular}{c|c|c|c|c}
        \toprule
        Stage & Operator  & Stride & \#Channels & \#Layers \\
        \midrule
        0  &    Conv3x3             & 2   &  24 & 1 \\
        1 &   Fused-MBConv1, k3x3   & 1   &  24   & 2  \\
        2 &   Fused-MBConv4, k3x3   & 2   &  48   & 4  \\
        3 &   Fused-MBConv4, k3x3   & 2   &  64   & 4  \\
        4 &   MBConv4, k3x3, SE0.25 & 2   &  128  & 6  \\
        5 &   MBConv6, k3x3, SE0.25 & 1   &  160  & 9  \\
        6 &   MBConv6, k3x3, SE0.25 & 2   &  256  & 15  \\
        7 &   Conv1x1 \& Pooling \& FC    & - & 1280 & 1 \\
        \bottomrule
    \end{tabular}
    }
\label{tab:arch}                                                    
\end{table} 
\paragraph{{\xnet} Scaling:}
We scale up {\xnet}-S to obtain {\xnet}-M/L using similar compound scaling as~\cite{efficientnet19}, with a few additional optimizations: (1) we restrict the maximum inference image size to 480, as very large images often lead to expensive memory and training speed overhead; (2) as a heuristic, we also gradually add more layers to later stages (e.g., stage 5 and 6 in Table \ref{tab:arch}) in order to increase the network capacity without adding much runtime overhead.

\begin{figure}[!ht]
    \centering
    \includegraphics[width=0.9\columnwidth]{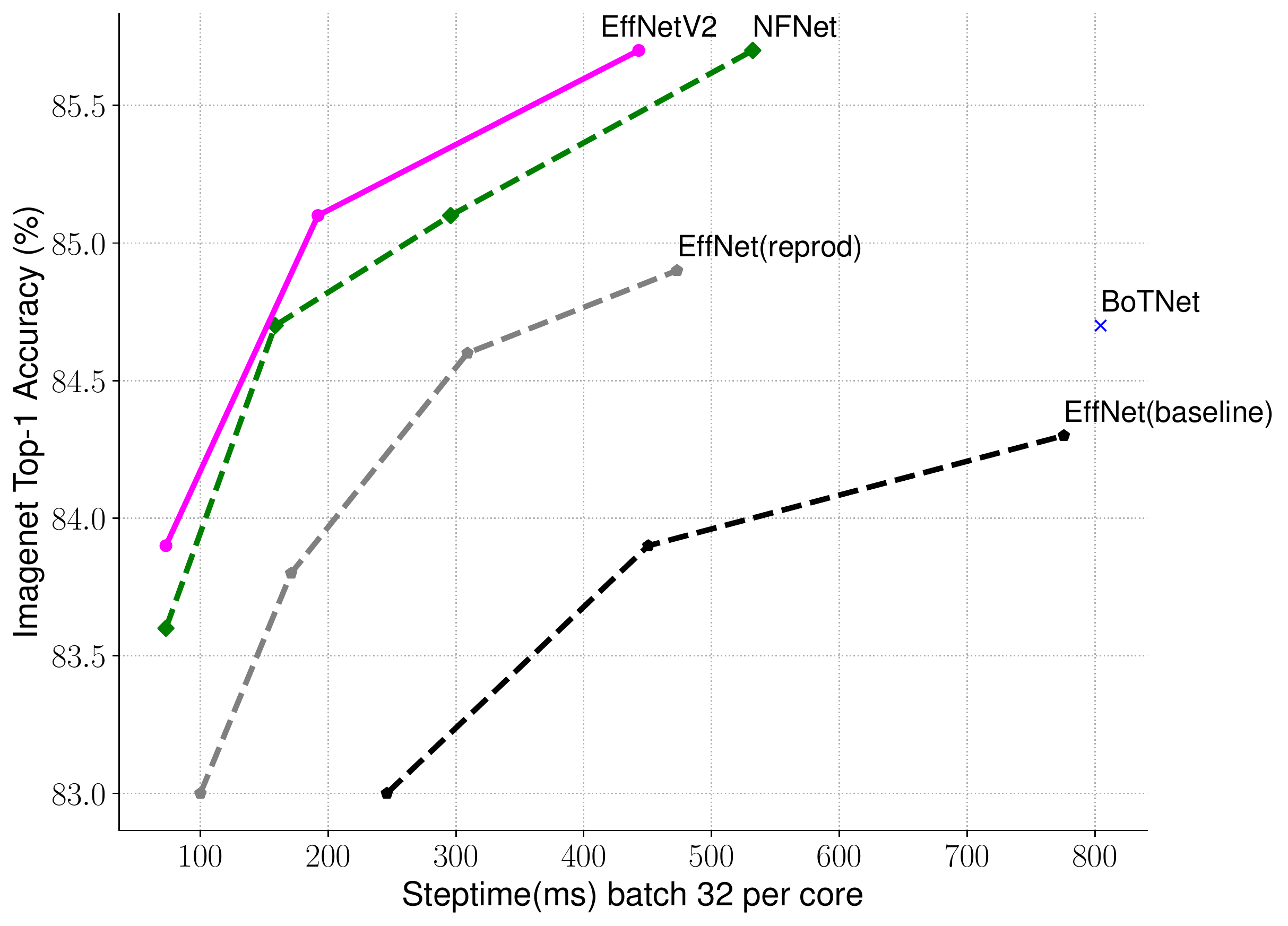}
    \vskip -0.15in
    \caption{ImageNet accuracy and training step time on TPUv3 -- Lower step time is better; all models are trained with fixed image size without progressive learning.}
     \label{fig:steptime}
    \vskip -0.1in
\end{figure}
 
\paragraph{Training Speed Comparison:}
Figure \ref{fig:steptime} compares the training step time for our new {\xnet}, where all models are trained with fixed image size without progressive learning. For EfficientNet~\cite{efficientnet19}, we show two curves: one is trained with the original inference size, and the other is trained with about 30\% smaller image size, same as {\xnet} and NFNet~\cite{fixres20,nfnet21}. All models are trained with 350 epochs, except  NFNets are trained with 360 epochs, so all models have a similar number of training steps. Interestingly, we observe that when trained properly, EfficientNets still achieve pretty strong performance trade-off. More importantly, with our training-aware NAS and scaling, our  proposed {\xnet} model train much faster than the other recent models. These results  also align with our inference results as shown in Table \ref{tab:imagenet} and Figure \ref{fig:infer2acc}.

\section{Progressive Learning}
\label{sec:plearn}

\subsection{Motivation}
\label{sec:motivation}

As discussed in Section \ref{sec:xnet}, image size plays an important role in training efficiency. In addition to FixRes~\cite{fixres20}, many other works dynamically change image sizes during training~\cite{fastaidawnbench,mixmatch19}, but they often cause a drop in accuracy.

We hypothesize the accuracy drop comes from the unbalanced regularization: when training with different image sizes, we should also adjust the regularization strength accordingly (instead of using a fixed regularization as in previous works). In fact, it is common that large models require stronger regularization to combat overfitting: for example, EfficientNet-B7 uses larger dropout and stronger data augmentation than the B0. In this paper, we argue that even for the same network, smaller image size leads to smaller network capacity and thus needs weaker regularization; vice versa, larger image size leads to more computations with larger capacity,  and thus more vulnerable to
overfitting.

To validate our hypothesis, we train a model, sampled from our search space, with different image sizes and data augmentations (Table \ref{tab:isize}). When image size is small, it has the best accuracy with weak augmentation; but for larger images, it performs better with stronger augmentation. This insight motivates us to adaptively adjust regularization along with image size during training, leading to our improved method of progressive learning.

\begin{table}[!h]
    \vskip -0.15in
    \caption{ImageNet top-1 accuracy. We use RandAug~\cite{randaug20}, and report mean and stdev for 3 runs.}
    \centering
    \resizebox{0.98\columnwidth}{!}{
        \begin{tabular}{lccc}
        \toprule
                    & Size=128 & Size=192 & Size=300 \\
        \midrule
          RandAug magnitude=5  & \textbf{78.3 \std{0.16}} & 81.2 \std{0.06} &  82.5	\std{0.05} \\
          RandAug magnitude=10  & 78.0 \std{0.08}         & \textbf{81.6 \std{0.08}} &  82.7 \std{0.08}  \\
          RandAug magnitude=15  & 77.7 \std{0.15}         & 81.5 \std{0.05} &  \textbf{83.2	\std{0.09}}  \\
        \bottomrule
        \end{tabular}
    }
    \label{tab:isize}
    \vskip -0.1in
\end{table} 

\subsection{\normalsize Progressive Learning with adaptive Regularization}

\begin{figure}[t]
    \centering
    \includegraphics[width=0.9\linewidth]{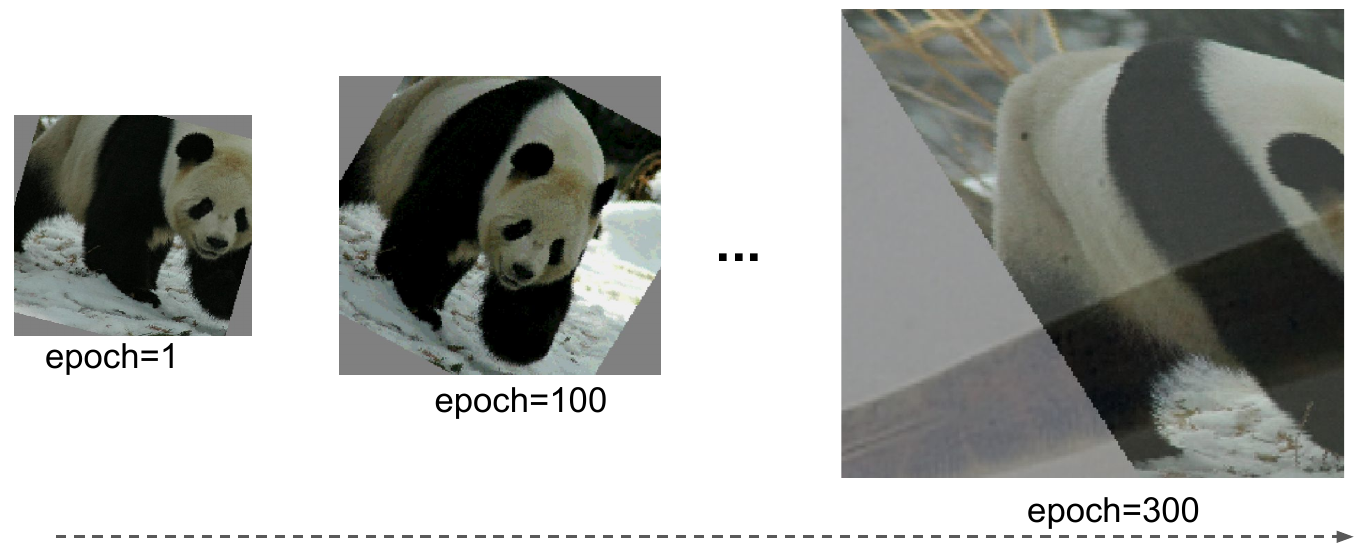}
    \vskip -0.1in
    \caption{
        Training process in our improved progressive learning -- It starts with small image size and weak regularization (epoch=1), and then gradually increase the learning difficulty with larger image sizes and stronger regularization: larger dropout rate, RandAugment magnitude, and mixup ratio (e.g., epoch=300).
    }
    \label{fig:ptrain}
    \vskip -0.1in
\end{figure}

Figure \ref{fig:ptrain} illustrates the training process of our improved progressive learning: in the early training epochs, we train the network with smaller images and weak regularization, such that the network can learn simple representations easily and fast. Then, we gradually increase image size but also making learning more difficult by adding stronger regularization.  Our approach is built upon ~\cite{fastaidawnbench} that progressively changes image size, but here we adaptively adjust regularization as well.

Formally, suppose the whole training has $N$ total steps, the target image size is $S_e$, with a list of regularization magnitude $\Phi_e = \{\phi^k_e\}$, where $k$ represents a type of regularization such as dropout rate or mixup rate value. We divide the training into $M$ stages: for each stage $1 \le i \le M$, the model is trained with image size $S_i$ and regularization magnitude $\Phi_i = \{\phi^k_i\}$. The last stage $M$ would use the targeted image size $S_e$ and regularization $\Phi_e$. For simplicity, we heuristically pick the initial image size $S_0$ and regularization $\Phi_0$, and then use a linear interpolation to determine the value for each stage. Algorithm \ref{alg:ptrain} summarizes the procedure. At the beginning of each stage, the network will inherit all weights from the previous stage. Unlike transformers, whose weights (e.g., position embedding) may depend on input length, ConvNet weights are independent to image sizes and thus can be inherited easily.

 \begin{algorithm}[!htb]
   \small
    \caption{\small Progressive learning with adaptive regularization.}
    \label{alg:ptrain}
 \begin{algorithmic}
    \STATE {\bfseries Input:} Initial image size $S_0$ and regularization $\{\phi^k_0\}$.
    \STATE {\bfseries Input:} Final image size $S_e$ and regularization $\{\phi^k_e\}$.
    \STATE {\bfseries Input:} Number of total training steps $N$ and stages $M$.
    \FOR{$i=0$ {\bfseries to} $M-1$}
    \STATE Image size: $S_i \leftarrow S_0 + (S_e - S_0)\cdot \frac{i}{M - 1}$
    \STATE Regularization: $R_i \leftarrow \{\phi_i^k = \phi_0^k + (\phi_e^k - \phi_0^k)\cdot \frac{i}{M - 1} \}$
    \STATE Train the model for $\frac{N}{M}$ steps with $S_i$ and $R_i$.
    \ENDFOR
 \end{algorithmic}
 \end{algorithm}

Our improved progressive learning is generally compatible to existing regularization. For simplicity, this paper mainly studies the following three types of regularization:
\begin{itemize}
    \setlength\itemsep{0em}
    \item{\BF{Dropout}~\cite{dropout14}:} a network-level regularization, which reduces co-adaptation by randomly dropping channels. We will adjust the dropout rate $\gamma $.
    \item{\BF{RandAugment}~\cite{randaug20}:} a per-image data augmentation, with adjustable magnitude $\epsilon$. 
    \item{\BF{Mixup}~\cite{mixup18}:} a cross-image data augmentation. Given two images with labels ($x_i$, $y_i$) and ($x_j$, $y_j$), it combines them with mixup ratio $\lambda$: $\tilde{x_i} = \lambda x_j + (1 - \lambda)x_i$ and $\tilde{y_i} = \lambda y_j + (1 - \lambda)y_i$. We would adjust mixup ratio $\lambda$ during training.
\end{itemize}

\section{Main Results}
\label{sec:results}

This section presents our experimental setups, the main results on ImageNet, and the transfer learning results on CIFAR-10, CIFAR-100, Cars, and Flowers.

\subsection{ImageNet ILSVRC2012}
\label{sec:imagenet1k}

\paragraph{Setup:}
ImageNet ILSVRC2012 \cite{imagenet15} contains about 1.28M training images and 50,000 validation images with 1000 classes.  During architecture search or hyperparameter tuning, we reserve 25,000 images (about 2\%) from the training set as \TT{minival} for accuracy evaluation. We also use \TT{minival} to perform early stopping. Our ImageNet training settings largely follow EfficientNets~\cite{efficientnet19}: RMSProp optimizer with decay 0.9 and momentum 0.9; batch norm momentum 0.99; weight decay 1e-5. Each model is trained for 350 epochs with total batch size 4096. Learning rate is first warmed up from 0 to 0.256, and then decayed by 0.97 every 2.4 epochs. We use exponential moving average with 0.9999 decay rate, RandAugment \cite{randaug20}, Mixup~\cite{mixup18}, Dropout~\cite{dropout14}, and stochastic depth \cite{droppath16} with 0.8 survival probability.

\begin{table}[!h]
    \vskip -0.1in
    \caption{
        Progressive training settings for {\xnet}.
       }
    \centering
    \resizebox{0.9\columnwidth}{!}{
        \begin{tabular}{l||cc|cc|cc}
        \toprule
                      &  \multicolumn{2}{c|}{S} & \multicolumn{2}{c|}{M}& \multicolumn{2}{c}{L} \\
                      &  min& max& min&max & min&max \\
        \midrule
          Image Size  &   128&300&  128&380& 128&380   \\
          RandAugment  &   5&15&   5&20&   5&25  \\
          Mixup alpha &   0&0&  0&0.2&  0&0.4 \\
          Dropout rate &  0.1&0.3&  0.1&0.4&  0.1&0.5 \\
        \bottomrule
        \end{tabular}
    }
    \label{tab:ptrain}
    \vskip -0.1in
\end{table} %
\begin{table*}
    \vskip -0.1in
    \caption{
        \textbf{{\xnet} Performance Results on ImageNet}~\cite{imagenet15} -- \TT{Infer-time} is measured on V100 GPU FP16 with batch size 16 using the same codebase \cite{pytorchimagemodel}; \TT{Train-time} is the total training time normalized for 32 TPU cores. Models marked with 21k are pretrained on ImageNet21k with 13M images, and others are directly trained on ImageNet ILSVRC2012 with 1.28M images from scratch. All {\xnet} models are trained with our improved method of progressive learning. 
       }
    \centering
    \resizebox{0.98\linewidth}{!}{
        \begin{tabular}{cl||rrrrr}
        \toprule [0.15em]
        & Model              &  Top-1 Acc. & Params & FLOPs & Infer-time(ms)   &  Train-time (hours) \\
        \midrule
        \multirow{22}{*}{\shortstack{ConvNets \\ \& Hybrid}}
          &EfficientNet-B3 \cite{efficientnet19}&   81.5\%  &     12M   &   1.9B  &    19     &  10 \\
          &EfficientNet-B4 \cite{efficientnet19}&   82.9\%  &     19M   &   4.2B  &    30     &  21 \\
          &EfficientNet-B5 \cite{efficientnet19}&   83.7\%  &     30M   &   10B   &    60     &  43 \\
          &EfficientNet-B6 \cite{efficientnet19}&   84.3\%  &     43M   &   19B   &    97    &  75 \\
          &EfficientNet-B7 \cite{efficientnet19}&   84.7\%  &     66M   &   38B   &    170    & 139 \\
          &RegNetY-8GF   \cite{regnet20}        &   81.7\%  &     39M   &   8B   &     21    & -    \\
          &RegNetY-16GF   \cite{regnet20}       &   82.9\%  &     84M   &   16B   &     32    & -    \\
          &ResNeSt-101  \cite{resnest20}        &   83.0\%  &     48M   &   13B   &     31    & -    \\
          &ResNeSt-200  \cite{resnest20}        &   83.9\%  &     70M   &   36B   &     76    & -    \\
          &ResNeSt-269  \cite{resnest20}        &   84.5\%  &    111M   &   78B   &    160    & -    \\
          &TResNet-L \cite{tresnet20}  &  83.8\%   &   56M     &   -   &    45     & -  \\
          &TResNet-XL \cite{tresnet20} &  84.3\%   &   78M     &   -   &    66    & -  \\

          &EfficientNet-X \cite{efficientnetx21} &  84.7\%   &   73M     &   91B   &    -     & -  \\

          &NFNet-F0 \cite{nfnet21}              &   83.6\%  &     72M   &   12B   &     30    & 8.9    \\ 
          &NFNet-F1 \cite{nfnet21}              &   84.7\%  &    133M   &   36B   &     70    & 20    \\ 
          &NFNet-F2 \cite{nfnet21}              &   85.1\%  &    194M   &   63B   &    124    & 36    \\ 
          &NFNet-F3 \cite{nfnet21}              &   85.7\%  &    255M   &  115B   &    203    & 65    \\ 
          &NFNet-F4 \cite{nfnet21}              &   85.9\%  &    316M   &  215B   &    309    & 126   \\ 
          &LambdaResNet-420-hybrid \cite{lambdanet21}& 84.9\% &   125M       &    -    &    -      & 67  \\
          &BotNet-T7-hybrid \cite{botnet21}            &  84.7\%   &   75M     &   46B   &    -      & 95  \\
          &BiT-M-R152x2 (21k)  \cite{bit20}     &   85.2\%  &    236M   &  135B   &    500    &  -   \\
        \midrule
        \multirow{8}{*}{\shortstack{Vision \\ Transformers}}
          &ViT-B/32 \cite{vit21}                &   73.4\%  &     88M   &   13B   &     13    & -    \\
          &ViT-B/16  \cite{vit21}               &   74.9\%  &     87M   &   56B   &     68    & -    \\
          &DeiT-B (ViT+reg)  \cite{deit21}               &   81.8\%  &     86M   &   18B   &     19    & -    \\
          &DeiT-B-384  (ViT+reg) \cite{deit21}            &   83.1\%  &     86M   &   56B   &    68    & -    \\
          &T2T-ViT-19  \cite{t2tvit21}               &   81.4\%  &     39M  &   8.4B   &     -    & -    \\
          &T2T-ViT-24  \cite{t2tvit21}               &   82.2\%  &     64M   &   13B   &     -    & -    \\
          &ViT-B/16 (21k) \cite{vit21}          &   84.6\%  &    87M   &  56B   &    68   &  - \\
          &ViT-L/16 (21k) \cite{vit21}          &   85.3\%  &    304M   &  192B   &    195   &  172 \\
        \midrule
        \multirow{6}{*}{\shortstack{ConvNets \\ (ours)}}
          &\BF{{\xnet}-S}                             &   83.9\%  &    22M    &   8.8B    &    24    &  7.1 \\
          &\BF{{\xnet}-M}                             &   85.1\%  &    54M    &    24B  &     57    &  13 \\
          &\BF{{\xnet}-L}                             &   85.7\%  &   120M    &    53B  &     98    &  24 \\
          &\BF{{\xnet}-S (21k)}                       &   84.9\% &    22M    &    8.8B  &     24    &  9.0 \\
          &\BF{{\xnet}-M (21k)}                       &   86.2\%  &   54M    &    24B  &     57    &  15 \\
          &\BF{{\xnet}-L (21k)}                       &   86.8\%  &   120M    &    53B  &     98    &  26 \\
          &\BF{{\xnet}-XL (21k)}                      &   87.3\%  &   208M    &    94B  &     -    &  45 \\
        \bottomrule[0.15em]
        \multicolumn{7}{l}{~~We do not include models pretrained on non-public Instagram/JFT images, or models with extra distillation or ensemble.~~}
        \end{tabular}
    }
    \label{tab:imagenet}
    \vskip -0.05in
\end{table*}
 
For progressive learning, we divide the training process into four stages with about 87 epochs per stage: the early stage uses a small image size with weak regularization, while the later stages use larger image sizes with stronger regularization, as described in Algorithm \ref{alg:ptrain}. Table \ref{tab:ptrain} shows the minimum (for the first stage) and maximum (for the last stage) values of image size and regularization. For simplicity, all models use the same minimum values of size and regularization, but they adopt different maximum values, as larger models generally require more regularization to combat overfitting. Following~\cite{fixefficientnet20}, our maximum image size for training is about 20\% smaller than inference, but we don't finetune any layers after training.

\begin{figure*}
    \centering
    \begin{subfigure}[t]{0.33\textwidth}
        \centering
        \includegraphics[width=1.0\columnwidth]{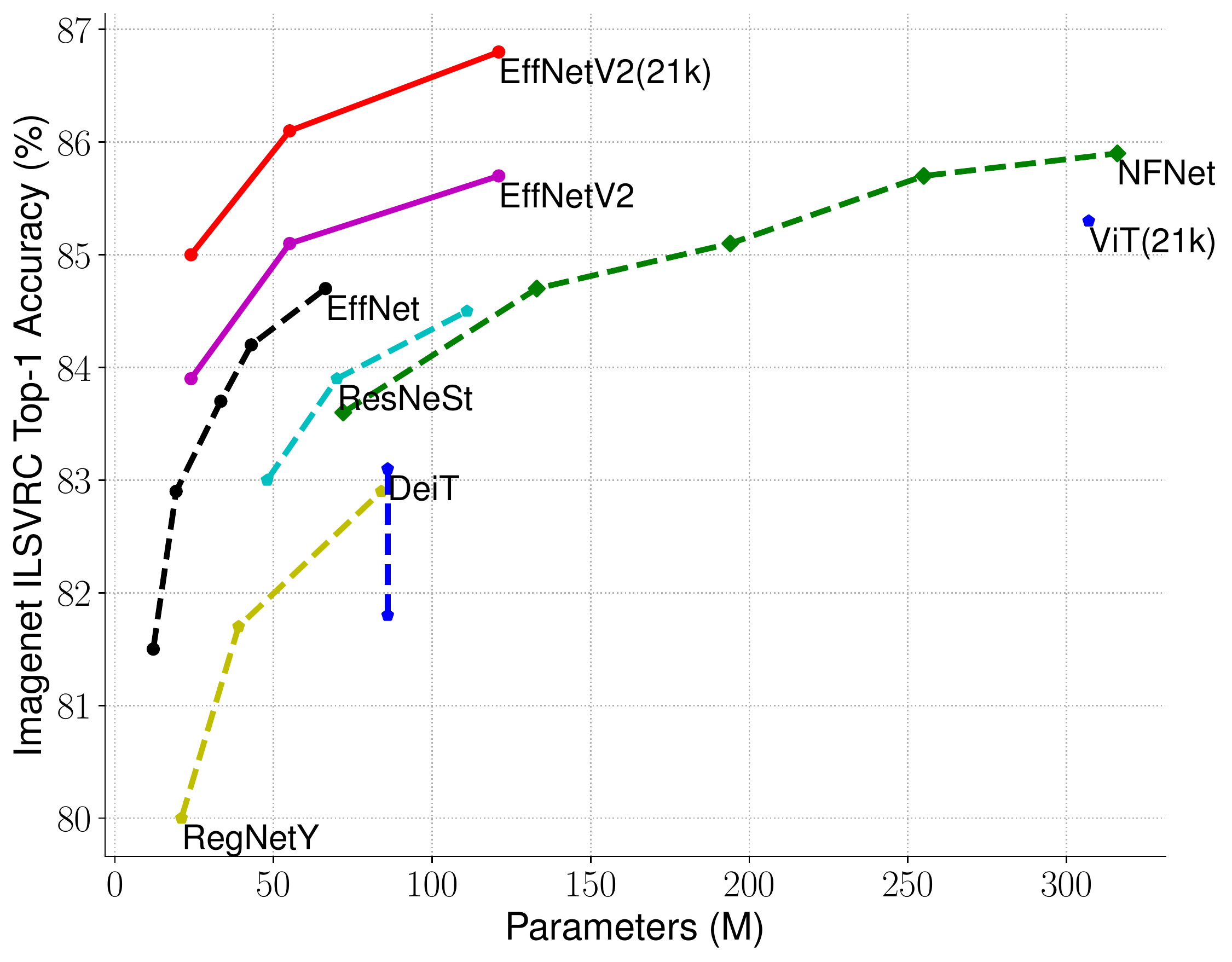}
        \caption{Parameters}
    \end{subfigure}
    \begin{subfigure}[t]{0.33\textwidth}
        \centering
		\includegraphics[width=1.0\columnwidth]{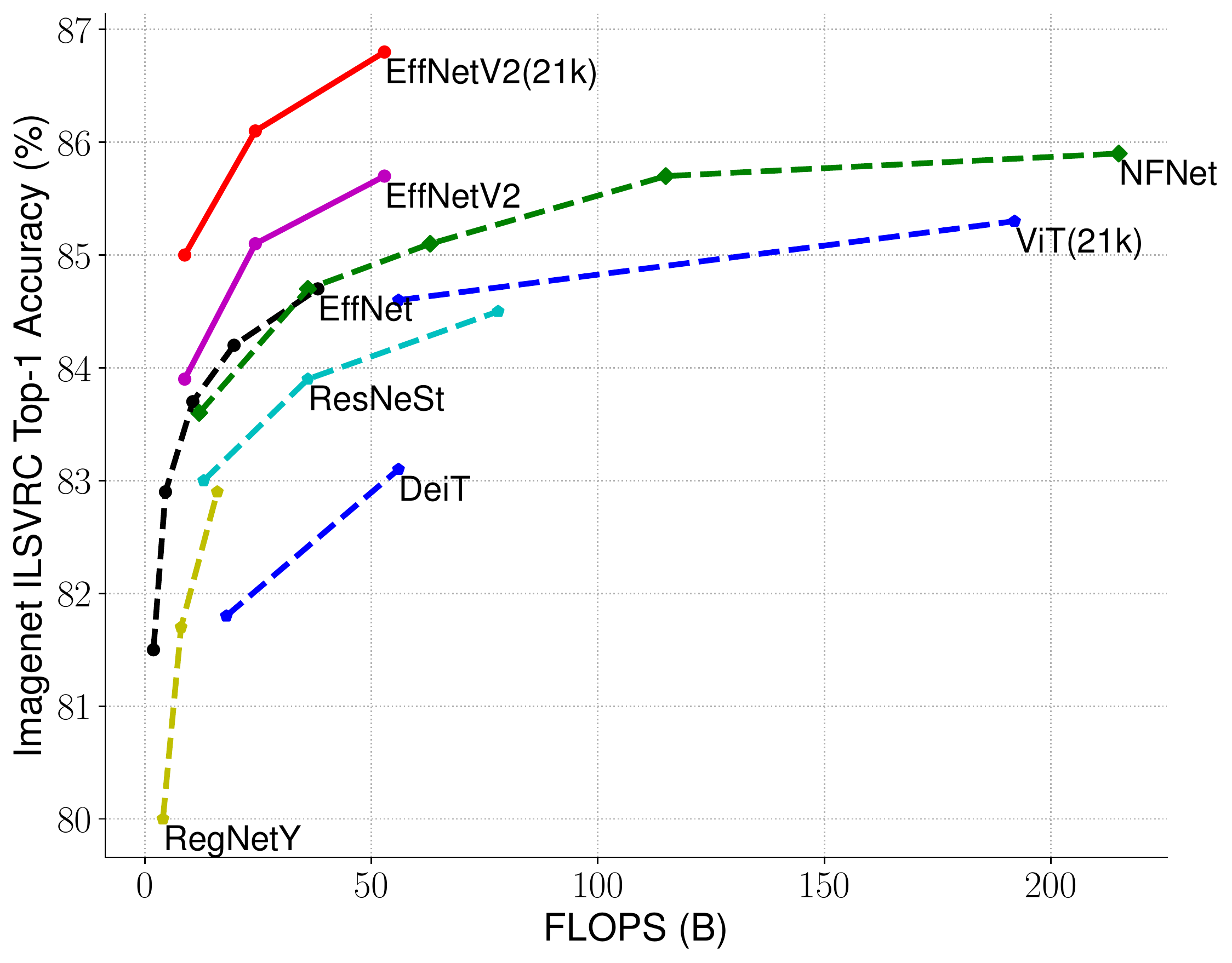}
		\caption{FLOPs}
    \end{subfigure}
    \begin{subfigure}[t]{0.33\textwidth}
		\centering
		\includegraphics[width=1.0\columnwidth]{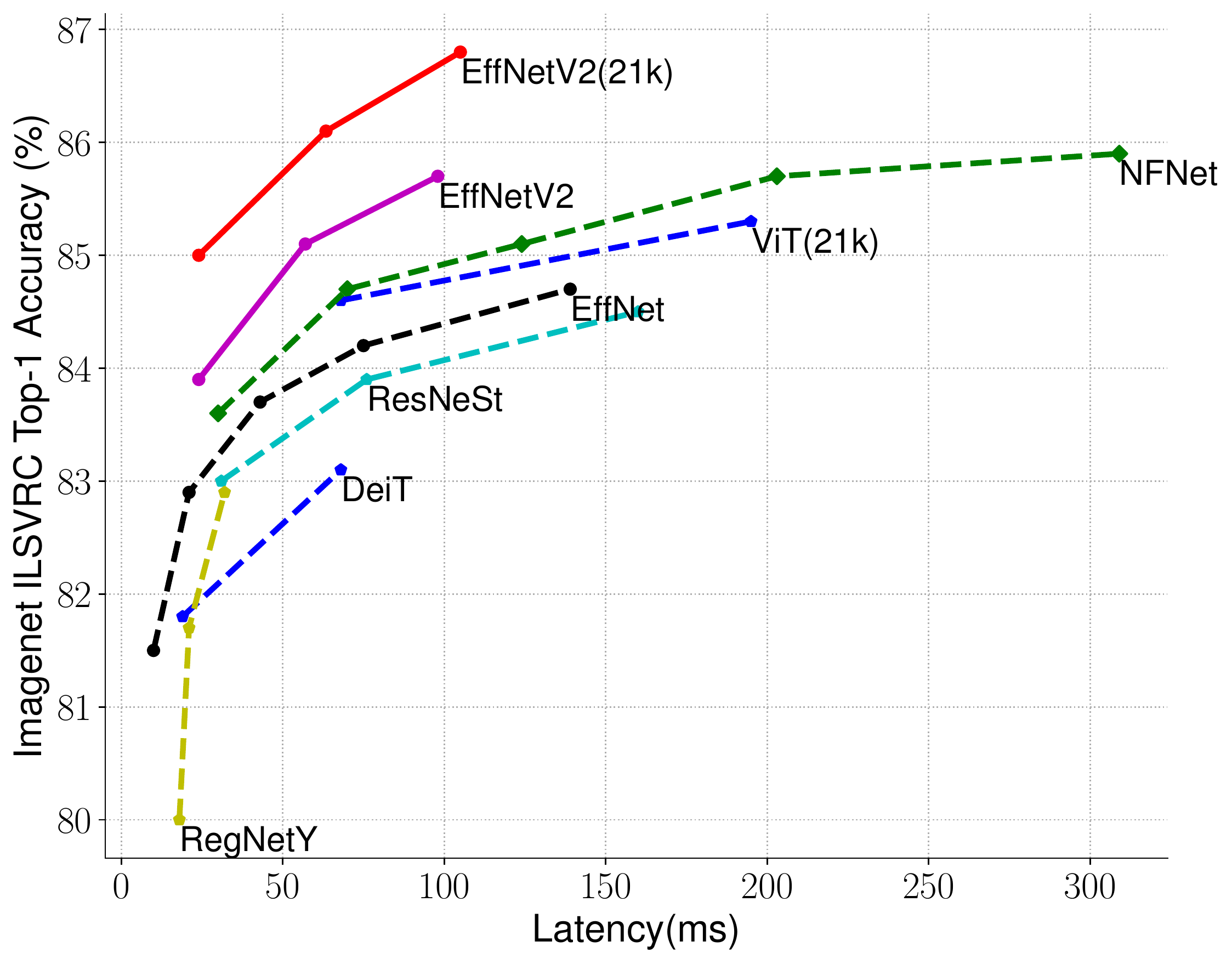}
		\caption{GPU V100 Latency (batch 16)}
	\end{subfigure}
	\vskip -0.1in
    \caption{
    	\BF{Model Size, FLOPs, and Inference Latency}  -- Latency is measured with batch size 16 on V100 GPU. \TT{21k} denotes pretrained on ImageNet21k images, others are just trained on ImageNet ILSVRC2012. Our {\xnet} has slightly better parameter efficiency with EfficientNet, but runs 3x faster for inference.
    }
     \label{fig:infer2acc}
\end{figure*}
 
\paragraph{Results:}
As shown in Table \ref{tab:imagenet}, our {\xnet} models are significantly faster and achieves better accuracy and parameter efficiency than previous ConvNets and Transformers on ImageNet. In particular, our {\xnet}-M achieves comparable accuracy to EfficientNet-B7 while training 11x faster using the same computing resources. Our {\xnet} models also significantly outperform all recent RegNet and ResNeSt, in both accuracy and inference speed. Figure~\ref{fig:train2acc} further visualizes the comparison on training speed and parameter efficiency. Notably, this speedup is a combination of progressive training and better networks, and we will study the individual impact for each of them in our ablation studies.

Recently, Vision Transformers have demonstrated impressive results on ImageNet accuracy and training speed. However, here we show that properly designed ConvNets with improved training method can still largely outperform vision transformers in both accuracy and training efficiency. In particular, our {\xnet}-L achieves 85.7\% top-1 accuracy, surpassing ViT-L/16(21k), a much larger transformer model pretrained on a larger ImageNet21k dataset. Here, ViTs are not well tuned on ImageNet ILSVRC2012; DeiTs use the same architectures as ViTs, but achieve better results by adding more regularization.

Although our {\xnet} models are optimized for training, they also perform well for inference, because training speed  often correlates with inference speed.  Figure \ref{fig:infer2acc} visualizes the model size, FLOPs, and inference latency based on Table \ref{tab:imagenet}. Since latency often depends on hardware and software, here we use the same PyTorch Image Models codebase~\cite{pytorchimagemodel} and run all models on the same machine using the batch size 16. In general, our models have slightly better parameters/FLOPs efficiency than EfficientNets, but our inference latency is up to 3x faster than EfficientNets. Compared to the recent ResNeSt that are specially optimized for GPUs, our {\xnet}-M achieves 0.6\% better accuracy with 2.8x faster inference speed.

\subsection{ImageNet21k}
\paragraph{Setup:} ImageNet21k~\cite{imagenet15} contains about 13M training images with 21,841 classes. The original ImageNet21k doesn't have train/eval split, so we reserve randomly picked 100,000 images as validation set and use the remaining as training set. We largely reuse the same training settings as ImageNet ILSVRC2012 with a few changes: (1) we change the training epochs to 60 or 30 to reduce training time, and use cosine learning rate decay that can adapt to different steps without extra tuning; (2) since each image has multiple labels, we normalize the labels to have sum of 1 before computing softmax loss. After pretrained on ImageNet21k, each model is finetuned on ILSVRC2012 for 15 epochs using cosine learning rate decay.

\begin{table*}
    \caption{
        \textbf{Transfer Learning Performance Comparison} -- All models are pretrained on ImageNet ILSVRC2012 and finetuned on downstream datasets. Transfer learning accuracy is averaged over five runs.
       }
    \centering
    \resizebox{0.95\linewidth}{!}{
        \begin{tabular}{cl||cc|llll}
        \toprule 
        &Model              &  Params  & ImageNet Acc. & CIFAR-10 & CIFAR-100 & Flowers  & Cars \\
        \midrule
        \multirow{2}{*}{ConvNets}
          &GPipe \cite{gpipe18}               &      556M     & 84.4  &    99.0 &  91.3  & 98.8  & 94.7 \\ 
          &EfficientNet-B7 \cite{efficientnet19} &      66M     & 84.7  &    98.9 &  91.7 &  98.8  & 94.7 \\ 
        \midrule
        \multirow{6}{*}{\shortstack{Vision\\ Transformers}}
          &ViT-B/32 \cite{vit21}         &      88M   &   73.4  &  97.8 &  86.3 &  85.4  & - \\
          &ViT-B/16  \cite{vit21}        &      87M   &   74.9   & 98.1 &  87.1 &  89.5  & - \\
          &ViT-L/32  \cite{vit21}        &     306M   &   71.2   & 97.9 &  87.1 &  86.4  & - \\
          &ViT-L/16  \cite{vit21}        &     306M   &   76.5   & 97.9 &  86.4 &  89.7  & - \\
          &DeiT-B  (ViT+regularization)  \cite{deit21}        &      86M   &   81.8   & 99.1 &  90.8 &  98.4  & 92.1 \\
          &DeiT-B-384  (ViT+regularization) \cite{deit21}     &      86M   &   83.1   & 99.1 &  90.8 &  98.5  & 93.3 \\
          \midrule
        \multirow{3}{*}{\shortstack{ConvNets \\ (ours)}}
          &{\xnet}-S                             &     24M    &   83.2 & 98.7\std{0.04} & 91.5\std{0.11}& 97.9\std{0.13} & 93.8\std{0.11} \\
          &{\xnet}-M                             &     55M    &  85.1 & 99.0\std{0.08} & 92.2\std{0.08}& 98.5\std{0.08} & 94.6\std{0.10}\\ 
          &{\xnet}-L                             &    121M    &  85.7 & \BF{99.1}\std{0.03} &  \BF{92.3}\std{0.13} & \BF{98.8}\std{0.05}  & \BF{95.1}\std{0.10}  \\
        \bottomrule[0.15em]
        \end{tabular}
    }
    \label{tab:transfer}
\end{table*} 
\paragraph{Results:}
Table \ref{tab:imagenet} shows the performance comparison, where models tagged with \TT{21k} are pretrained on ImageNet21k and finetuned on ImageNet ILSVRC2012. Compared to the recent ViT-L/16(21k), our {\xnet}-L(21k) improves the top-1 accuracy by 1.5\% (85.3\% vs. 86.8\%), using 2.5x fewer parameters and 3.6x fewer FLOPs, while running 6x - 7x faster in training and inference.

We would like to highlight a few interesting observations:
\begin{itemize}
    \setlength\itemsep{0em}
    \item \emph{Scaling up data size is more effective than simply scaling up model size in high-accuracy regime}: when the top-1 accuracy is beyond 85\%, it is very difficult to further improve it by simply increasing model size due to the severe overfitting. However, the extra ImageNet21K pretraining can  significantly improve accuracy.  The effectiveness of large datasets is also observed in previous works~\cite{imagenetinstagram18,noisystudent20,vit21}. %
    \item \emph{Pretraining on ImageNet21k could be quite efficient.} Although ImageNet21k has 10x more data, our training approach enables us to finish the pretraining of {\xnet} within two days using 32 TPU cores (instead of weeks for ViT~\cite{vit21}). This is more effective than training larger models on ImageNet. 
    We suggest future research on large-scale models use the public ImageNet21k as a default dataset.
\end{itemize}

\subsection{Transfer Learning Datasets}
\paragraph{Setup:} We evaluate our models on four transfer learning datasets: CIFAR-10, CIFAR-100, Flowers and Cars. Table \ref{tab:transferdata} includes the statistics of these datasets.

\begin{table}[!ht]
    \vskip -0.15in
    \caption{
        Transfer learning datasets.
       }
    \centering
    \resizebox{0.95\linewidth}{!}{
        \begin{tabular}{l|ccc}
        \toprule
                      &  Train images  & Eval images & Classes \\
        \midrule
        CIFAR-10 \cite{cifar} &  50,000 & 10,000 & 10 \\
        CIFAR-100 \cite{cifar} &  50,000 & 10,000 & 100  \\
        Flowers \cite{flowers} & 2,040 & 6,149 & 102 \\
        Cars \cite{stanfordcars} &  8,144 & 8,041 & 196 \\
        \bottomrule
        \end{tabular}
    }
    \label{tab:transferdata}
    \vskip -0.1in
\end{table} 

For this experiment, we use the checkpoints trained on ImageNet ILSVRC2012. For fair comparison, no ImageNet21k images are used here. Our finetuning settings are mostly the same as ImageNet training with a few modifications similar to \cite{vit21,deit21}: We use smaller batch size 512, smaller initial learning rate 0.001 with cosine decay. For all datasets, we train each model for fixed 10,000 steps. Since each model is finetuned with very few steps, we disable weight decay and use a simple cutout data augmentation.

\paragraph{Results:} Table \ref{tab:transfer} compares the transfer learning performance. In general, our models outperform previous ConvNets and Vision Transformers for all these datasets, sometimes by a non-trivial margin: for example, on CIFAR-100, {\xnet}-L achieves 0.6\% better accuracy than prior GPipe/EfficientNets and 1.5\% better accuracy than prior ViT/DeiT models. These results suggest that our models also generalize well beyond ImageNet.

\section{Ablation Studies}
\label{sec:ablation}

\subsection{Comparison to EfficientNet}
In this section, we will compare our {\xnet} (V2 for short) with EfficientNets~\cite{efficientnet19} (V1 for short) under the same training and inference settings. 

\paragraph{Performance with the same training:}  Table \ref{tab:arche2x} shows the performance comparison using the same progressive learning settings. As we apply the same progressive learning to EfficientNet, its training speed (reduced from 139h to 54h) and accuracy (improved from 84.7\% to 85.0\%) are better than the original paper~\cite{efficientnet19}. However, as shown in Table \ref{tab:arche2x}, our {\xnet} models still outperform EfficientNets by a large margin: {\xnet}-M reduces parameters by 17\% and FLOPs by 37\%, while running 4.1x faster in training and 3.1x faster in inference than EfficientNet-B7. Since we are using the same training settings here, we attribute the gains to the {\xnet} architecture.

\begin{table}[h]
    \vskip -0.1in
    \caption{
        Comparison with the same training settings -- 
        Our new {\xnet}-M runs faster with less parameters.
       }
    \centering
    \resizebox{0.95\columnwidth}{!}{
        \begin{tabular}{l||cllll}
        \toprule [0.15em]
                    & Acc.& Params & FLOPs & TrainTime & InferTime  \\
                    & (\%) & (M)  & (B)  & (h)  & (ms) \\
        \midrule
          V1-B7      & 85.0  & 66 & 38 & 54 & 170\\
          V2-M (ours) & 85.1  & 55 \good{-17\%} & 24 \good{-37\%} & 13 \good{-76\%} & \;\;57 \good{-66\%}\\
        \bottomrule[0.15em]
        \end{tabular}
    }
    \label{tab:arche2x}
    \vskip -0.1in
\end{table} 
\paragraph{Scaling Down:} Previous sections mostly focus on large-scale models. Here we compare smaller models by scaling down our {\xnet}-S using EfficientNet compound scaling. For easy comparison, all models are trained without progressive learning. Compared to small-size EfficientNets (V1), our new {\xnet} (V2) models are generally faster while maintaining comparable parameter efficiency.

\begin{table}[h]
    \vskip -0.15in
    \centering
    \caption{
    Scaling down model size -- We measure the inference throughput (images/sec) on V100 FP16 GPU with batch size 128.}
    \resizebox{0.9\columnwidth}{!}{
     \begin{tabular}{lrrrr}
        \toprule
                       & Top-1 Acc.  & Parameters & FLOPs & Throughput  \\
        \midrule
        V1-B1 & 79.0\%  &  7.8M    &  0.7B & 2675\\
        V2-B0 & 78.7\%  &  7.4M    &  0.7B & \good{2.1x} 5739\\
        \hline
        V1-B2 & 79.8\%  &  9.1M   & 1.0B & 2003 \\
        V2-B1 & 79.8\%  &  8.1M   & 1.2B &  \good{2.0x} 3983 \\
        \hline
        V1-B4  & 82.9\%  &  19M   & 4.2B & 628 \\
        V2-B3 & 82.1\%  &  14M   & 3.0B &  \good{2.7x} 1693 \\
        \hline
        V1-B5 & 83.7\%  &  30M    & 9.9B & 291 \\
        V2-S  & 83.6\%  &  24M    & 8.8B & \good{3.1x} 901  \\
        \bottomrule
     \end{tabular}
    }
    \vskip -0.1in
\label{tab:v1v2compare}
\end{table} 
\subsection{Progressive Learning for Different Networks}
We ablate the performance of our progressive learning for different networks. Table \ref{tab:resnet} shows the performance comparison between our progressive training and the baseline training,  using the same ResNet and EfficientNet models.  Here, the baseline ResNets have higher accuracy than the original paper~\cite{resnet16} because they are trained with our improved training settings (see Section \ref{sec:results}) using more epochs and better optimizers. We also increase the image size from 224 to 380 for ResNets to further increase the network capacity and accuracy. 

\begin{table}[!h]
    \vskip -0.1in
    \centering
    \caption{
        Progressive learning for ResNets and EfficientNets -- (224) and (380) denote inference image size. Our progressive training improves both accuracy and training time for all networks.
       }
    \resizebox{0.99\columnwidth}{!}{
        \begin{tabular}{l||rr|rr}
        \toprule
                    & \multicolumn{2}{c|}{Baseline} &   \multicolumn{2}{c}{Progressive}  \\
                    & Acc.(\%) & TrainTime &   Acc.(\%) & TrainTime  \\
        \midrule
          ResNet50 (224)  & 78.1 &  4.9h  &  78.4   & 3.5h \good{-29\%}  \\
          ResNet50 (380)  & 80.0 &  14.3h  &  80.3    & 5.8h \good{-59\%}  \\
          ResNet152 (380) & 82.4 &  15.5h  &  82.9    & 7.2h \good{-54\%} \\
        \midrule
          EfficientNet-B4 & 82.9 &  20.8h  &  83.1    & 9.4h \good{-55\%} \\
          EfficientNet-B5 & 83.7 &  42.9h  &  84.0    & 15.2h \good{-65\%}\\
        \bottomrule
        \end{tabular}
    }
    \label{tab:resnet}
\end{table} 
As shown in Table \ref{tab:resnet}, our progressive learning generally reduces the training time and meanwhile improves the accuracy for all different networks. Not surprisingly, when the default image size is very small, such as ResNet50(224) with 224x224  size, the training speedup is limited (1.4x speedup); however, when the default image size is larger and the model is more complex, our approach achieves larger gains on accuracy and training efficiency: for ResNet152(380), our approach improves speed up the training by 2.1x with slightly better accuracy;  for EfficientNet-B4, our approach improves speed up the training by 2.2x.

\subsection{Importance of Adaptive Regularization}

A key insight from our training approach is the adaptive regularization, which dynamically adjusts regularization according to image size. This paper chooses a simple progressive approach for its simplicity, but it is also a general method that can be combined with other approaches.

\begin{figure}[h]
    \vskip -0.1in
    \centering
    \includegraphics[width=0.7\linewidth]{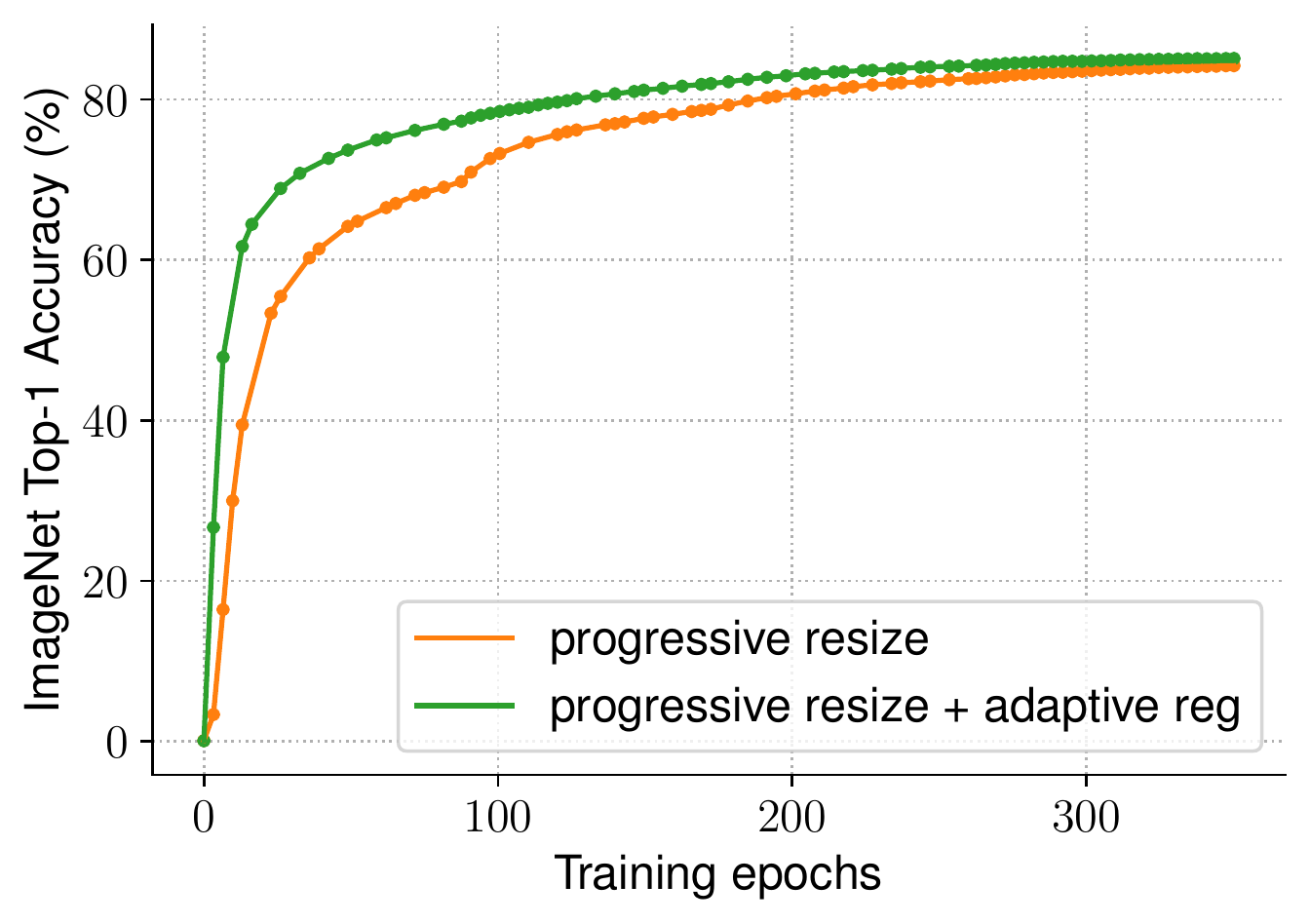}
    \vskip -0.2in
    \caption{
        \BF{Training curve comparison} -- Our adaptive regularization converges faster and achieves better final accuracy.
    }
    \label{fig:ptraincurve}
    \vskip -0.1in
\end{figure}

Table \ref{tab:reg} studies our adaptive regularization on two training settings: one is to progressively increase image size from small to large~\cite{fastaidawnbench}, and the other is to randomly sample a different image size for each batch~\cite{mixmatch19}. Because TPU needs to recompile the graph for each new size, here we randomly sample a image size every eight epochs instead of every batch. Compared to the vanilla approaches of progressive or random resizing that use the same regularization for all image sizes, our adaptive regularization improves the accuracy by 0.7\%. Figure \ref{fig:ptraincurve} further compares the training curve for the progressive approach. Our adaptive regularization uses much smaller regularization for small images at the early training epochs, allowing models to converge faster and achieve better final accuracy.

\begin{table}[!h]
    \vskip -0.1in
    \caption{
         Adaptive regularization -- We compare ImageNet top-1 accuracy based on the average of three runs.
       }
    \centering
    \resizebox{0.99\columnwidth}{!}{
        \begin{tabular}{ccc}
        \toprule
                 &      Vanilla  &   +our adaptive reg  \\
        \midrule
        Progressive resize \cite{fastaidawnbench}  & 84.3\std{0.14}   &  85.1\std{0.07} \good{+0.8} \\
        \hline
        Random resize  \cite{mixmatch19}   &  83.5\std{0.11}   &  84.2\std{0.10} \good{+0.7} \\
        \bottomrule
        \end{tabular}
    }
    \label{tab:reg}
    \vskip -0.1in
\end{table} 

\section{Conclusion}
\label{sec:conclude}
This paper presents {\xnet}, a new family of smaller and faster neural networks for image recognition. Optimized with training-aware NAS and model scaling, our {\xnet} significantly outperforms previous models, while being much faster and more efficient in parameters. To further speed up the training, we propose an improved method of progressive learning, that jointly increases image size and regularization during training. Extensive experiments show our {\xnet} achieves strong results on ImageNet, and CIFAR/Flowers/Cars. Compared to EfficientNet and more recent works, our {\xnet}  trains up to 11x faster while being up to 6.8x smaller. \section*{Acknowledgements} Special thanks to Lucas Sloan for helping open sourcing. We thank Ruoming Pang, Sheng Li, Andrew Li, Hanxiao Liu, Zihang Dai, Neil Houlsby, Ross Wightman, Jeremy Howard, Thang Luong, Daiyi Peng, Yifeng Lu, Da Huang, Chen Liang, Aravind Srinivas, Irwan Bello,  Max Moroz, Futang Peng for their feedback.

\bibliographystyle{sty/icml2021}
\bibliography{cv}

\end{document}